\documentclass[letterpaper, 10 pt, conference]{ieeeconf}  

\IEEEoverridecommandlockouts

\usepackage[utf8]{inputenc}
\usepackage[T1]{fontenc}
\usepackage{hyperref}
\usepackage{url}
\usepackage{booktabs}
\usepackage{amsfonts}
\usepackage{nicefrac}
\usepackage{microtype}
\usepackage{tablefootnote}

\hypersetup{
    colorlinks=true,
    linkcolor=blue,
    filecolor=magenta,      
    urlcolor=cyan,
}
\usepackage[dvipsnames,table]{xcolor}
\usepackage{graphicx}
\usepackage{bm}
\usepackage{amsmath,amssymb}
\usepackage{xspace}

\usepackage{nicefrac}
\usepackage{tabulary,multirow,overpic}
\usepackage{subfig}
\usepackage[normalem]{ulem}

\newcolumntype{x}[1]{>{\centering\arraybackslash}p{#1pt}}
\newcommand{\tablestyle}[2]{\setlength{\tabcolsep}{#1}\renewcommand{\arraystretch}{#2}\centering\footnotesize}
\newlength\savewidth\newcommand\shline{\noalign{\global\savewidth\arrayrulewidth
  \global\arrayrulewidth 1pt}\hline\noalign{\global\arrayrulewidth\savewidth}}

\newcommand*{\eg}{\textit{e.g.}\@\xspace}
\newcommand*{\ie}{\textit{i.e.}\@\xspace}

\definecolor{demphcolor}{RGB}{100,100,100}

\newcommand{\app}{\raise.17ex\hbox{$\scriptstyle\sim$}}
\definecolor{lightgray}{RGB}{230,230,230}

\def\methodname{YolactEdge}
\def\featflow{FeatFlowNet}

\title{\LARGE \bf \methodname{}: Real-time Instance Segmentation on the Edge}

\author{Haotian Liu$^{*}$, Rafael A. Rivera Soto$^{*}$, Fanyi Xiao, and Yong Jae Lee
\thanks{$^{1}$Fanyi Xiao is with Amazon Web Services, Inc., the rest are with the University of California, Davis. {\tt\small \{lhtliu, riverasoto, fyxiao, yongjaelee\}@ucdavis.edu}
(*~\emph{Haotian Liu and Rafael A. Rivera Soto are co-first authors.})}
}

\begin{document}

\maketitle
\thispagestyle{empty}
\pagestyle{empty}

\begin{abstract}

We propose \emph{YolactEdge}, the first competitive instance segmentation approach that runs on small edge devices at real-time speeds. Specifically, YolactEdge runs at up to 30.8 FPS on a Jetson AGX Xavier (and 172.7 FPS on an RTX 2080 Ti) with a ResNet-101 backbone on 550x550 resolution images.  To achieve this, we make two improvements to the state-of-the-art image-based real-time method YOLACT~\cite{bolya2019yolact}: (1) applying TensorRT optimization while carefully trading off speed and accuracy, and (2) a novel feature warping module to exploit temporal redundancy in videos.  Experiments on the YouTube VIS and MS COCO datasets demonstrate that YolactEdge produces a 3-5x speed up over existing real-time methods while producing competitive mask and box detection accuracy. We also conduct ablation studies to dissect our design choices and modules. Code and models are available at \url{https://github.com/haotian-liu/yolact_edge}.
\vspace{-1pt}
\end{abstract}

\section{Introduction}

Instance segmentation is a challenging problem that requires the correct detection and segmentation of each \emph{object instance} in an image. A fast and accurate instance segmenter would have many useful applications in robotics, autonomous driving, image/video retrieval, healthcare, security, and others.  In particular, a \emph{real-time} instance segmenter that can operate on \emph{small edge devices} is necessary for many real-world scenarios. For example, in safety critical applications in complex environments, robots, drones, and other autonomous machines may need to perceive objects and humans in real-time on device -- without having access to the cloud, and in resource constrained settings where bulky and power hungry GPUs (e.g., Titan Xp) are impractical.  However, while there has been great progress in real-time instance segmentation research~\cite{bolya2019yolact,lee2019centermask,wang2020solov2,zhang2020mask,chen2020blendmask,yolact-plus-tpami2020,peng2020deep}, thus far, there is no method that can run accurately at real-time speeds on small edge devices like the Jetson AGX Xavier.  

In this paper, we present \emph{\methodname{}}, a novel real-time instance segmentation approach that runs accurately on edge devices at real-time speeds.  Specifically, with a ResNet-101 backbone, \methodname{} runs at up to 30.8 FPS on a Jetson AGX Xavier (and 172.7 FPS on an RTX 2080 Ti GPU), which is 3-5x faster than existing state-of-the-art real-time methods, while being competitive in accuracy.

In order to perform inference at real-time speeds on edge devices, we build upon the state-of-the-art image-based real-time instance segmentation method, YOLACT~\cite{bolya2019yolact}, and make two fundamental improvements, one at the system-level and the other at the algorithm-level: (1) we apply NVIDIA's TensorRT inference engine~\cite{TensorRT} to quantize the network parameters to fewer bits while systematically balancing any tradeoff in accuracy, and (2) we leverage temporal redundancy in video (i.e., temporally nearby frames are highly correlated), and learn to transform and propagate features over time so that the deep network's expensive backbone feature computation does not need to be fully computed on every frame. 

The proposed shift to video from static image processing makes sense from a practical standpoint, as the real-time aspect matters much more for video applications that require low latency and real-time response than for image applications; \eg, for real-time control in robotics and autonomous driving, or real-time object/activity detection in security and augmented reality, where the system must process a stream of video frames and generate instance segmentation outputs in real-time.  Importantly, all existing real-time instance segmentation methods (including YOLACT) are static image-based, which makes \methodname{} the first \emph{video-dedicated} real-time instance segmentation method.    

In sum, our contributions are: (1) we apply TensorRT optimization while carefully trading off speed and accuracy, (2) we propose a novel feature warping module to exploit temporal redundancy in videos, (3) we perform experiments on the benchmark image MS COCO~\cite{lin2014microsoft} and video YouTube VIS~\cite{yang2019video} datasets, demonstrating a 3-5x faster speed compared to existing real-time instance segmentation methods while being competitive in accuracy, and (4) we publicly release our code and models to facilitate progress in robotics applications that require on device real-time instance segmentation.

\section{Related Work}

\textbf{Real-time instance segmentation in images.} YOLACT~\cite{bolya2019yolact} is the first real-time instance segmentation method to achieve competitive accuracy on the challenging MS COCO \cite{lin2014microsoft} dataset. Recently, CenterMask~\cite{lee2019centermask}, BlendMask~\cite{chen2020blendmask}, and SOLOv2~\cite{wang2020solov2} have improved accuracy in part by leveraging more accurate object detectors (\eg, FCOS~\cite{tian2019fcos}). All existing real-time instance segmentation approaches~\cite{bolya2019yolact,lee2019centermask,chen2020blendmask,yolact-plus-tpami2020,wang2020solov2} are image-based and require bulky GPUs like the Titan Xp / RTX 2080 Ti to achieve real-time speeds. In contrast, we propose the first \emph{video-based} real-time instance segmentation approach that can run on small edge devices like the Jetson AGX Xavier.

\textbf{Feature propagation in videos} has been used to improve speed and accuracy for video classification and video object detection~\cite{zhu2017deep,zhu2017flow,zhu2018towards}. These methods use off-the-shelf optical flow networks~\cite{dosovitskiy2015flownet} to estimate pixel-level object motion and warp feature maps from frame to frame. However, even the most lightweight flow networks \cite{dosovitskiy2015flownet,sun2018pwc} require non-negligible memory and compute, which are obstacles for real-time speeds on edge devices. In contrast, our model estimates object motion and performs feature warping directly at the feature level (as opposed to the input pixel level), which enables real-time speeds.

\textbf{Improving model efficiency.} Designing lightweight yet performant backbones and feature pyramids has been one of the main thrusts in improving deep network efficiency. MobileNetv2~\cite{sandler2018mobilenetv2} introduces depth-wise convolutions and inverted residuals to design a lightweight architecture for mobile devices. MobileNetv3~\cite{howard2019mobilenetv3}, NAS-FPN~\cite{ghiasi2019nasfpn}, and EfficientNet~\cite{tan2019efficientnet} use neural architecture search to automatically find efficient architectures. Others utilize knowledge distillation~\cite{hinton2015distillation,polino2018compression,Sanh2019DistilBERTAD}, model compression~\cite{han2015deep_compression,SqueezeNet}, or binary networks~\cite{rastegariECCV16,bulat-bmvc2019}. The CVPR Low Power Computer Vision Challenge participants have used TensorRT~\cite{TensorRT}, a deep learning inference optimizer, to quantize and speed up object detectors such as Faster-RCNN on the NVIDIA Jetson TX2~\cite{alyamkin2019lowpower}. In contrast to most of these approaches, \methodname{} retains large expressive backbones, and exploits temporal redundancy in video together with a TensorRT optimization for fast and accurate instance segmentation.  

\section{Approach}

Our goal is to create an instance segmentation model, \methodname{}, that can achieve real-time ($>$30 FPS) speeds on edge devices.  To this end, we make two improvements to the image-based real-time instance segmentation approach YOLACT~\cite{bolya2019yolact}: (1) applying TensorRT optimization, and (2) exploiting temporal redundancy in video.

\subsection{TensorRT Optimization}

The edge device that we develop our model on is the NVIDIA Jetson AGX Xavier. The Xavier is equipped with an integrated Volta GPU with Tensor Cores, dual deep learning accelerator, 32GB of memory, and reaches up to 32 TeraOPS at a cost of \$699. Importantly, the Xavier is the only architecture from the NVIDIA Jetson series that supports both FP16 and INT8 Tensor Cores, which are needed for TensorRT~\cite{tensorrt_tensorcore} optimization.

TensorRT is NVIDIA's deep learning inference optimizer that provides mixed-precision support, optimal tensor layout, fusing of network layers, and kernel specializations~\cite{TensorRT}. A major component of accelerating models using TensorRT is the quantization of model weights to INT8 or FP16 precision.  Since FP16 has a wider range of precision than INT8, it yields better accuracy at the cost of more computational time. Given that the weights of different deep network components (backbone, prediction module, etc.) have different ranges, this speed-accuracy trade-off varies from component to component. Therefore, we convert each model component to TensorRT independently and explore the optimal mix between INT8 and FP16 weights that maximizes FPS while preserving accuracy.

\begin{table}[t]
  \centering
  \scriptsize
  \vspace{8pt}
  \begin{tabular}{ccccccc}
    \shline
    Backbone & FPN & ProtoNet & PredHead & TensorRT & mAP & FPS \\
    \hline
    FP32 & FP32 & FP32 & FP32 & N & \textbf{29.8}  & 6.4 \\
    FP16 & FP16 & FP16 & FP16 & N & 29.7 & 12.1 \\
    \hline
    FP32 & FP32 & FP32 & FP32 & Y & 29.6 & 19.1 \\
    FP16 & FP16 & FP16 & FP16 & Y & 29.7 & 21.9 \\
    \hline
    INT8 & FP16 & FP16 & FP16 & Y & 29.9 & 26.3 \\
    INT8 & FP16 & INT8 & FP16 & Y & 29.9 & 26.5 \\
    \rowcolor{lightgray} \textbf{INT8} & \textbf{INT8} & \textbf{FP16} & \textbf{FP16} & Y & 29.7 & \textbf{27.7} \\
    INT8 & INT8 & INT8 & FP16 & Y & 29.8 & 27.4 \\
    \hline
    INT8 & FP16 & FP16 & INT8 & Y & 25.4 & 26.2 \\
    INT8 & FP16 & INT8 & INT8 & Y & 25.4 & 25.9 \\
    INT8 & INT8 & FP16 & INT8 & Y & 25.2 & 26.9 \\
    INT8 & INT8 & INT8 & INT8 & Y & 25.2 & 26.5 \\
    \shline
  \end{tabular}
  \caption{\textbf{Effect of Mixed Precision} on YOLACT~\cite{bolya2019yolact} with a ResNet-101 backbone on the MS COCO val2017 dataset with a Jetson AGX Xavier using 100 calibration images. Mixing precision across the modules results in different instance segmentation mean Average Precision (mAP) and FPS for each instantiation of YOLACT. All results are averaged over 5 runs, with a standard deviation less than 0.6 FPS.
  }
  \label{table:tensorrt_exhaustive}
\end{table}

\begin{figure*}[t!]
    \centering
    \includegraphics[width=\textwidth]{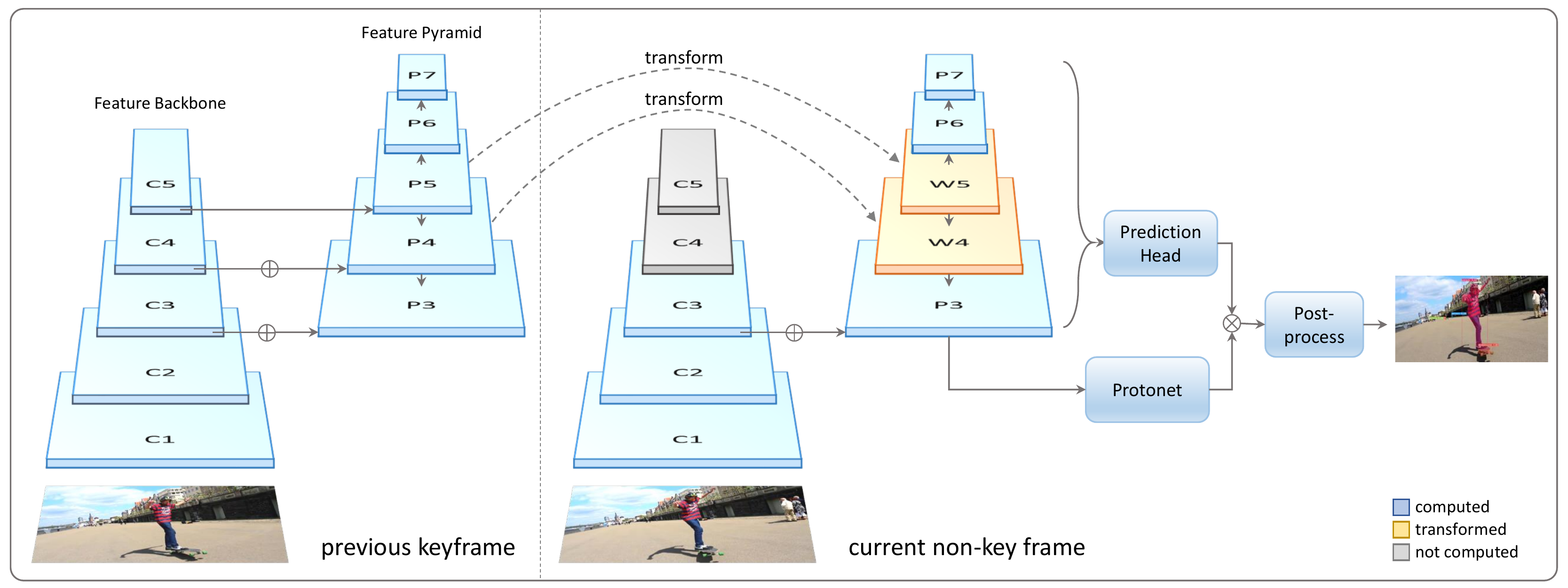}
    \caption{\textbf{\methodname} extends YOLACT~\cite{bolya2019yolact} to video by transforming a subset of the features from keyframes (left) to non-keyframes (right), to reduce expensive backbone computation. 
    Specifically, on non-keyframes, we compute $C_3$ features that are cheap while crucial for mask prediction given its high-resolution. This largely accelerates our method while retaining accuracy on non-keyframes.
    We use {\color{blue} blue}, {\color{orange} orange}, and {\color{gray} grey} to indicate computed, transformed, and skipped blocks, respectively. }
    \label{fig:yolact}
\end{figure*}

Table~\ref{table:tensorrt_exhaustive} shows this analysis for YOLACT~\cite{bolya2019yolact}, which is the baseline model that YolactEdge directly builds upon. Briefly, YOLACT can be divided into 4 components: (1) a feature backbone, (2) a feature pyramid network~\cite{lin2017feature} (FPN), (3) a ProtoNet, and (4) a Prediction Head; see Fig.~\ref{fig:yolact} (right) for the network architecture.  (More details on YOLACT will be provided in Sec.~\ref{sec:video}.)  The second row in Table~\ref{table:tensorrt_exhaustive} represents YOLACT, with all components in FP32 (i.e., no TensorRT optimization), and results in only 6.6 FPS on the Jetson AGX Xavier with a ResNet-101 backbone. From there, INT8 or FP16 conversion on different model components leads to various improvements in speed and changes in accuracy. Notably, conversion of the Prediction Head to INT8 (last four rows) always results in a large loss of instance segmentation accuracy. We hypothesize that this is because the final box and mask predictions require more than $2^8=256$ bins to be encoded without loss in the final representation.  Converting every component to INT8 except for the Prediction Head and FPN (row highlighted in gray) achieves the highest FPS with little mAP degradation. Thus, this is the final configuration we go with for our model in our experiments, but different configurations can easily be chosen based on need. 

In order to quantize model components to INT8 precision, a calibration step is necessary: TensorRT collects histograms of activations for each layer, generates several quantized distributions with different thresholds, and compares each of them to the reference distribution using KL Divergence~\cite{tensorrtcalibration}. This step ensures that the model loses as little performance as possible when converted to INT8 precision. Table \ref{ablation:int8-calibration} shows the effect of the calibration dataset size. We observe that calibration is necessary for accuracy, and generally a larger calibration set provides a better speed-accuracy trade-off.

\subsection{Exploiting Temporal Redundancy in Video}
\label{sec:video}

The TensorRT optimization leads to a $\sim$4x improvement in speed, and when dealing with static images, this is the version of YolactEdge one should use.  However, when dealing with \emph{video}, we can exploit temporal redundancy to make YolactEdge even faster, as we describe next.  

Given an input video as a sequence of frames $\{I_i\}$, we aim to predict masks for each object instance in each frame $\{y_i = \mathcal{N}(I_i)\}$, in a fast and accurate manner.  For our video instance segmentation network $\mathcal{N}$, we largely follow the YOLACT~\cite{bolya2019yolact} design for its simplicity and impressive speed-accuracy tradeoff. Specifically, on each frame, we perform two parallel tasks: (1) generating a set of prototype masks, and (2) predicting per-instance mask coefficients. Then, the final masks are assembled through linearly combining the prototypes with the mask coefficients.

For clarity of presentation, we decompose $\mathcal{N}$ into $\mathcal{N}_{feat}$ and $\mathcal{N}_{pred}$, where $\mathcal{N}_{feat}$ denotes the feature backbone stage and $\mathcal{N}_{pred}$ is the rest (\ie, prediction heads for class, box,  mask coefficients, and ProtoNet for generating prototype masks) which takes the output of $\mathcal{N}_{feat}$ and make instance segmentation predictions. We selectively divide frames in a video into two groups: keyframes $I^k$ and non-keyframes $I^n$; the behavior of our model on these two groups of frames only varies in the backbone stage.
\begin{align}
  y^k &= \mathcal{N}_{pred}(\mathcal{N}_{feat}(I^{k})) \\
  y^n &= \mathcal{N}_{pred}(\widetilde{\mathcal{N}}_{feat}(I^n))
\end{align}
For keyframes $I^k$, our model computes all backbone and pyramid features ($C_1-C_5$ and $P_3-P_7$ in Fig.~\ref{fig:yolact}). Whereas for non-keyframes $I^n$, we compute only a subset of the features, and transform the rest from the temporally closest previous keyframe using the mechanism that we elaborate on next. This way, we strike a balance between producing accurate predictions while maintaining a fast runtime. 

\vspace{5pt}
\noindent\textbf{Partial Feature Transform.} Transforming (\ie, warping) features from neighboring keyframes was shown to be an effective strategy for reducing backbone computation to yield fast video bounding box object detectors in~\cite{zhu2017deep}. Specifically, \cite{zhu2017deep} transforms all the backbone features using an off-the-shelf optical flow network~\cite{dosovitskiy2015flownet}.  However, due to inevitable errors in optical flow estimation, we find that it fails to provide sufficiently accurate features required for pixel-level tasks like instance segmentation. In this work, we propose to perform \textit{partial feature transforms} to improve the quality of the transformed features while still maintaining a fast runtime.

\begin{figure*}[t!]
    \centering
    \includegraphics[width=.8\textwidth]{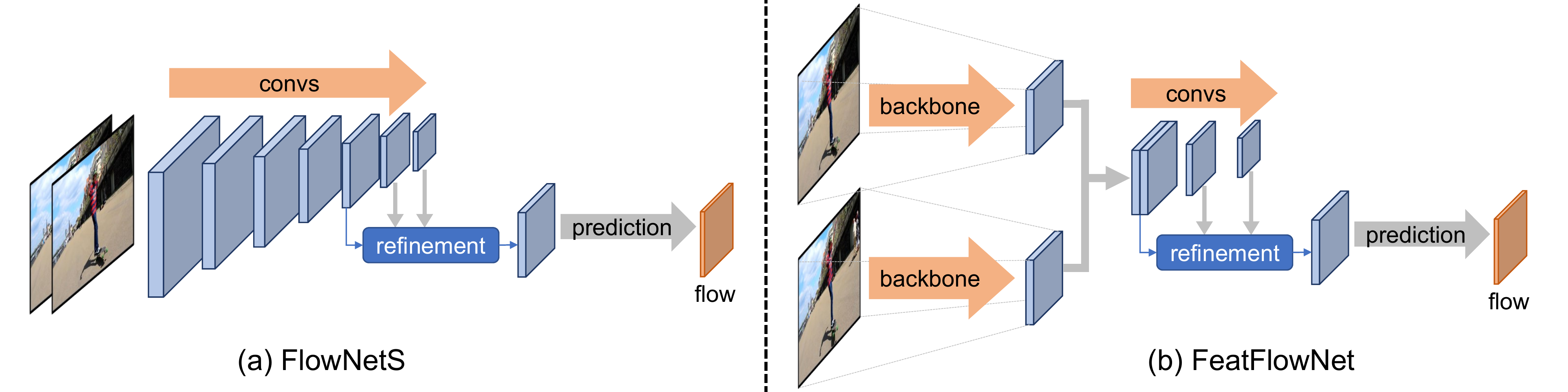}
    \caption{\textbf{Flow estimation}. Illustration of the difference between FlowNetS~\cite{dosovitskiy2015flownet} (a) and our \featflow{} (b).}
    \label{fig:flow-estimation}
\end{figure*}

Specifically, unlike~\cite{zhu2017deep}, which transforms all features ($P^k_3$, $P^k_4$, $P^k_5$ in our case) from a keyframe $I^k$ to a non-keyframe $I^n$, our method computes the backbone features for a non-keyframe only up through the high-resolution $C^n_3$ level (i.e., skipping $C^n_4$, $C^n_5$ and consequently $P^n_4$, $P^n_5$ computation), and only transforms the lower resolution $P^k_4$/$P^k_5$ features from the previous keyframe to approximate $P^n_4$/$P^n_5$ (denoted as $W^n_4$/$W^n_5$) in the current non-keyframe, as shown in Fig.~\ref{fig:yolact} (right).  It computes $P^n_6$/$P^n_7$ by downsampling $W^n_5$ in the same way as YOLACT. With the computed $C^n_3$ features and transformed $W^n_4$ features, it then generates $P^n_3$ as $P^n_3 = C^n_3 + up(W^n_4)$, where $up(\cdot)$ denotes upsampling. Finally, we use the $P^n_3$ features to generate pixel-accurate prototypes. This way, in contrast to~\cite{zhu2017deep}, we can preserve high-resolution details for generating the mask prototypes, as the high-resolution $C_3$ features are computed instead of transformed and thus are immune to errors in flow estimation.

Importantly, although we compute the $C_1$-$C_3$ backbone features for every frame (\ie, both key and non-keyframes), we avoid computing the most expensive part of the backbone, as the computational costs in different stages of pyramid-like networks are highly imbalanced. As shown in Table~\ref{table:compute-stat}, more than 66\% of the computation cost of ResNet-101 lies in $C_4$, while more than half of the inference time is occupied by backbone computation. By computing only lower layers of the feature pyramid and transforming the rest, we can largely accelerate our method to reach real-time performance.

\begin{table}[t!]
  \centering
  \footnotesize
  \subfloat[\textbf{ResNet-101 Backbone}]{
    \label{table:resnet-cost}
    \tabcolsep=0.01cm
    \begin{tabular}{c|x{20} x{20} x{20} x{20} x{20}}
      \shline
       & $C_1$ & $C_2$ & $C_3$ & $C_4$ & $C_5$ \\
      \hline
      \# of convs & 1 & 9 & 12 & \textbf{69} & 9\\
      TFLOPS & 0.1 & 0.7 & 1.0 & \textbf{5.2} & 0.8 \\
      \multicolumn{1}{c|}{\%} & 1.5 & 8.7 & 13.2 & \textbf{66.2} & 10.3 \\
      \shline
    \end{tabular}
  }
  \subfloat[\textbf{YOLACT}]{
    \label{table:yolact-cost}
    \tabcolsep=0.03cm
    \begin{tabular}{c|x{20}|c|x{20}}
      \shline
      Stage & \% & Stage & \% \\
      \hline
      Backbone & \textbf{54.7} & FPN & 6.4 \\
      ProtoNet & 7.8 & Pred & 10.6 \\
        Detect & 6.6 & Other & 13.1 \\
      \shline
    \end{tabular}
  }
  \caption{\textbf{Computational cost breakdown} for different stages of (a) ResNet-101 backbone, and (b) YOLACT.}
  \label{table:compute-stat}
\end{table}

In summary, our \textit{partial feature transform} design produces higher quality feature maps that are required for instance segmentation, while also enabling real-time speeds.

\vspace{5pt}
\noindent\textbf{Efficient Motion Estimation.} In this section, we describe how we efficiently compute flow between a keyframe and non-keyframe.  Given a non-keyframe $I^n$ and its preceding keyframe $I^k$, our model first encodes object motion between them as a 2-D flow field $\mathcal{M}(I^k, I^n)$.  It then uses the flow field to transform the features $F^k = \{P_4^k, P_5^k\}$ from frame $I^k$ to align with frame $I^n$ to produce the warped features $\widetilde{F}^{n} = \{W_4^n, W_5^n\} = \mathcal{T} (F^k, \mathcal{M}(I^k, I^n))$.

In order to perform fast feature transformation, we need to estimate object motion efficiently. Existing frameworks~\cite{zhu2017deep,zhu2017flow} that perform flow-guided feature transform directly adopt off-the-shelf pixel-level optical flow networks for motion estimation. FlowNetS~\cite{dosovitskiy2015flownet} (Fig.~\ref{fig:flow-estimation}a), for example, performs flow estimation in three stages: it first takes in raw RGB frames as input and computes a stack of features; it then refines a subset of the features by recursively upsampling and concatenating feature maps to generate coarse-to-fine features that carry both high-level (large motion) and fine local information (small motion); finally, it uses those features to predict the final flow map.

In our case, to save computation costs, instead of taking an off-the-shelf flow network that processes raw RGB frames, we reuse the features computed by our model's backbone network, which already produces a set of semantically rich features. To this end, we propose \featflow{} (Fig.~\ref{fig:flow-estimation}b), which generally follows the FlowNetS architecture, but in the first stage, instead of computing feature stacks \emph{from raw RGB image inputs}, we re-use features from the ResNet backbone ($C_3$) and use fewer convolution layers.  As we demonstrate in our experiments, our flow estimation network is much faster while being equally effective.

\vspace{5pt}
\noindent\textbf{Feature Warping.} We use \featflow{} to estimate the flow map ${\mathcal{M}}(I^k, I^n)$ between the previous keyframe $I^k$ and the current non-keyframe $I^n$, and then transform the features from $I^k$ to $I^n$ via inverse warping: by projecting each pixel $x$ in $I^n$ to $I^k$ as $x + \delta x$, where $\delta x = {\mathcal{M}_x}(I^k, I^n)$. The pixel value is then computed via bilinear interpolation $F^{k \to n}(x) = \sum_{u} \theta(u, x + \delta x) F^k(x)$, where $\theta$ is the bilinear interpolation weight at different spatial locations.

\vspace{5pt}
\noindent\textbf{Loss Functions.} For the instance segmentation task, we use the same losses as YOLACT~\cite{bolya2019yolact} to train our model: classification loss $L_{cls}$, box regression loss $L_{box}$, mask loss $L_{mask}$, and auxiliary semantic segmentation loss $L_{aux}$.
For flow estimation network pre-training, like \cite{dosovitskiy2015flownet}, we use the endpoint error (EPE).

\section{Results}

In this section, we analyze \methodname{}'s instance segmentation accuracy and speed on the Jetson AGX Xavier and RTX 2080 Ti.  We compare to state-of-the-art real-time instance segmentation methods, and perform ablation studies to dissect our various design choices and modules.  

\begin{figure}[t!]
    \centering
    \includegraphics[width=.49\textwidth]{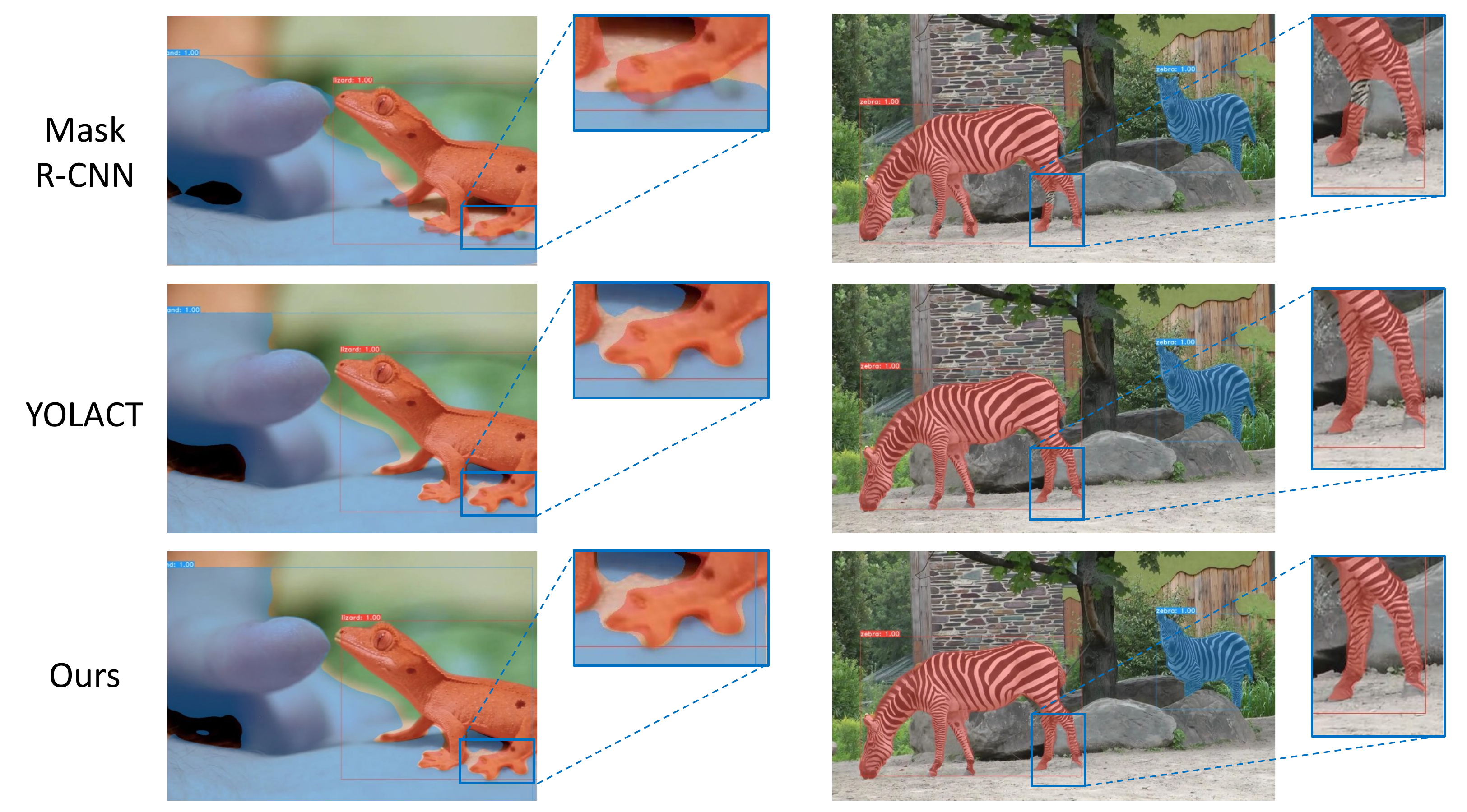}
    \caption{\textbf{Mask quality}. Our masks are as high quality as YOLACT even on non-keyframes, and are typically higher quality than those of Mask R-CNN \cite{he2017mask}.}
    \label{fig:mask-quality}
\end{figure}

\begin{figure*}[t!]
    \centering
    \includegraphics[width=\textwidth]{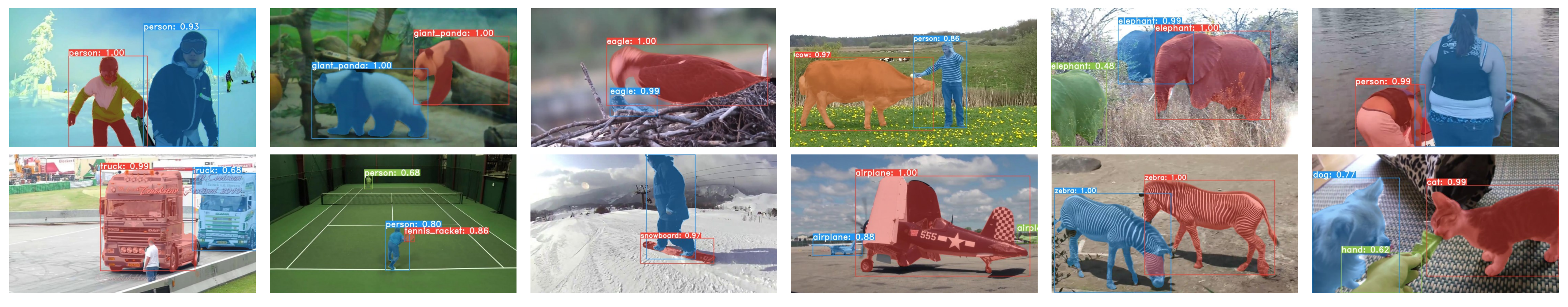}
    \caption{\textbf{\methodname{} results on YouTube VIS} on non-keyframes whose subset of features are warped from a keyframe 4 frames away (farthest in sampling window). Our mask predictions can tightly fit the objects, due to partial feature transform.}
    \label{fig:qualitative-results}
\end{figure*}

\vspace{5pt}
\noindent\textbf{Implementation details.} We train with a batch size of 32 on 4 GPUs using ImageNet pre-trained weights. We leave the pre-trained batchnorm (\textit{bn}) unfrozen and do not add any extra bn layers.  We first pre-train YOLACT with SGD for 500k iterations with $5 \times 10^{-4}$ initial learning rate. Then, we freeze YOLACT weights, and train \featflow{} on FlyingChairs \cite{DFIB15} with $2 \times 10^{-4}$ initial learning rate. Finally, we fine-tune all weights except ResNet backbone  for 200k iterations with $2 \times 10^{-4}$ initial learning rate. When pre-training YOLACT, we apply all data augmentations used in YOLACT; during fine-tuning, we disable \textit{random expand} to allow the warping module to model larger motions. For all training stages, we use cosine learning rate decay schedule, with weight decay $5 \times 10^{-4}$, and momentum $0.9$. We pick the first of every 5 frames as the keyframes. We use 100 images from the training set to calibrate our INT8 model components (backbone, prototype, \featflow{}) for TensorRT, and the remaining components (prediction head, FPN) are converted to FP16. We do not convert the warping module to TensorRT, as the conversion of the sampling function (needed for inverse warp) is not natively supported, and is also not a bottleneck for our feature propagation to be fast. We limit the output resolution to be a maximum of 640x480 while preserving the aspect ratio.

\begin{table}
  \footnotesize
  \centering
  \resizebox{0.49\textwidth}{!}{
  \begin{tabular}{l l c c c}
     \shline
    Method & Backbone & mask AP & box AP & RTX FPS \\ 
    \hline
    Mask R-CNN \cite{he2017mask} & R-101-FPN & 43.1 & 47.3 & 14.1 \\
    \hline
    CenterMask-Lite \cite{lee2019centermask} & V-39-FPN & 41.6 & 45.9 & 34.4 \\
    BlendMask-RT \cite{chen2020blendmask} & R-50-FPN & 44.0 & 47.9 & 49.3 \\
    SOLOv2-Light \cite{wang2020solov2} & R-50-FPN & 46.3 & -- & 43.9 \\
    YOLACT \cite{bolya2019yolact} & R-50-FPN & 44.7 & 46.2 & 59.8 \\
    YOLACT \cite{bolya2019yolact} & R-101-FPN & \textbf{47.3} & \textbf{48.9} & 42.6 \\
    \hline
    \textbf{Ours} \\
    \methodname{} (w/o TRT) & R-50-FPN & 44.2 & 45.2 & \textbf{67.0} \\
    \methodname{} (w/o TRT) & R-101-FPN & 46.9 & 47.8 & 61.2 \\
    \hline
    \methodname{} & R-50-FPN & 44.0 & 45.1 & \textbf{177.6} \\
    \methodname{} & R-101-FPN & 46.2 & 47.1 & 172.7 \\
    \shline
  \end{tabular}
  }
  \caption{\textbf{Comparison to state-of-the-art real-time methods on YouTube VIS}. We use our sub-training and sub-validation splits for YouTube VIS and perform joint training with COCO using a 1:1 data sampling ratio. (Box AP is not evaluated in the authors' code base of SOLOv2.)
  }
  \label{table:mask-results}
\end{table}

\begin{table}
\tabcolsep=0.1cm
\footnotesize
\resizebox{0.49\textwidth}{!}{
\centering
\begin{tabular}{ l l c c c c}
    \shline
    Method & Backbone & mask AP & box AP & AGX FPS & RTX FPS \\
    \hline
    YOLACT~\cite{bolya2019yolact} & MobileNet-V2 & \textbf{22.1} & \textbf{23.3} & 15.0 & 35.7 \\
    \methodname{} (w/o video) & MobileNet-V2 & 20.8 & 22.7 & \textbf{35.7}  & \textbf{161.4} \\
    \hline
    YOLACT~\cite{bolya2019yolact} & R-50-FPN & \textbf{28.2} & \textbf{30.3} & 9.1 & 45.0 \\
    \methodname{} (w/o video) & R-50-FPN & 27.0 & 30.1 & \textbf{30.7} & \textbf{140.3} \\
    \hline
    YOLACT~\cite{bolya2019yolact} & R-101-FPN & \textbf{29.8} & \textbf{32.3} & 6.6 & 36.5 \\
\methodname{} (w/o video) & R-101-FPN & 29.5 & 32.1 & \textbf{27.3} & \textbf{124.8} \\
    \shline
\end{tabular}}
\caption{\textbf{\methodname{} (w/o video) comparision to YOLACT on MS COCO~\cite{lin2014microsoft} test-dev split.} AGX: Jetson AGX Xavier; RTX: RTX 2080 Ti.}
\label{table:coco_results}
\end{table}

\begin{table}
\tabcolsep=0.1cm
\resizebox{0.49\textwidth}{!}{
\centering
\begin{tabular}{ l l x{35}x{35}x{35}x{35} }
    \shline
    Method & Backbone & mask AP & box AP & AGX FPS & RTX FPS \\
    \hline
    YOLACT~\cite{bolya2019yolact} & R-50-FPN & \textbf{44.7} & \textbf{46.2} & 8.5 & 59.8 \\
    \methodname{} (w/o TRT) & R-50-FPN & 44.2 & 45.2 & 10.5 & 67.0 \\
    \methodname{} (w/o video) & R-50-FPN & 44.5 & 46.0 & 32.0 & \textbf{185.7} \\
    \methodname{} & R-50-FPN & 44.0 & 45.1 & \textbf{32.4} & 177.6 \\
    \hline
    YOLACT~\cite{bolya2019yolact} & R-101-FPN & \textbf{47.3} & \textbf{48.9} & 5.9 & 42.6 \\
    \methodname{} (w/o TRT) & R-101-FPN & 46.9 & 47.8 & 9.5 & 61.2 \\
    \methodname{} (w/o video) & R-101-FPN & 46.9 & 48.4 & 27.9 & 158.2 \\
    \methodname{} & R-101-FPN & 46.2 & 47.1 & \textbf{30.8} & \textbf{172.7} \\
    \shline
\end{tabular}
}
\caption{\textbf{\methodname{} ablation results on Youtube VIS.}
\vspace{-2pt}
}
\label{table:yolactedge_ablation}
\end{table}

\begin{table*}[t!]
  \centering
  \footnotesize
  \captionsetup[subfloat]{captionskip=5pt}
  \captionsetup[subffloat]{justification=centering}
 \subfloat[\textbf{INT8 calibration} Effect of the number of calibration images.\label{ablation:int8-calibration}]{
  	\tablestyle{1.5pt}{1.05}
	\begin{tabular}{x{48} x{24} x{24}}
	  \shline
	  \#Calib. Img. & mAP & FPS \\
	  \hline
	  0 & 24.4 & -- \\
	  5 & 29.6 & 27.4 \\
	  50 & \textbf{29.8} & 27.4 \\
	  \rowcolor{lightgray} 100 & 29.7 & \textbf{27.5} \\
	  \shline
	\end{tabular}
  }\hspace{0.5em}
  \subfloat[\textbf{Partial feature transform} We warp $P_4$ \& $P_5$ as it is both fast and accurate. \label{ablation:partial-feature-transform}
  ]{
  	\tablestyle{1.5pt}{1.05}
    \begin{tabular}{r x{35} x{35}}
      \shline
      Warp layers & mAP & FPS \\ 
      \hline
      $C_4$, $C_5$ & \textbf{39.2} & 59.7 \\
      \rowcolor{lightgray} $P_4$, $P_5$ & \textbf{39.2} & 63.2 \\
      $C_3$, $C_4$, $C_5$ & 37.8 & 59.1 \\
      $P_3$, $P_4$, $P_5$ & 38.0 & \textbf{64.1} \\
      \shline
    \end{tabular}
  }\hspace{0.5em}
  \subfloat[\textbf{\featflow{}} We reduce channels for accuracy/speed tradeoff.
  \label{ablation:featflow-reduce-channel}
  ]{
  	\tablestyle{1.5pt}{1.05}
	\begin{tabular}{x{30} x{31} x{31}} 
	  \shline
	  Channels & mAP & FPS \\ 
	  \hline
	  \multicolumn{1}{c}{1x} & \textbf{47.0} & 48.3 \\
	  \multicolumn{1}{c}{1/2x} & 46.9 & 53.6 \\
	  \rowcolor{lightgray} \multicolumn{1}{c}{1/4x} & 46.9 & 61.2 \\
	  \multicolumn{1}{c}{1/8x} & -- & \textbf{62.2} \\
	  \shline
	\end{tabular}
  }\hspace{0.5em}
  \subfloat[\textbf{\featflow{}} is faster and equally effective compared to FlowNetS. \label{ablation:featflow-compare-flownets}
  ]{
  	\tablestyle{1.5pt}{1.05}
    \begin{tabular}{l x{30} x{30}}
      \shline
      Method & mAP & FPS \\ 
      \hline
      w/o flow & 31.8 & \textbf{72.5} \\
      FlowNetS & \textbf{39.2} & 43.3 \\
      \rowcolor{lightgray} \featflow{} & \textbf{39.2} & 61.2 \\
      \shline \\
    \end{tabular}
  }
  \vspace{-7pt}
  \caption{\textbf{Ablations}. (a) is on COCO val2017 using YOLACT with a R101 backbone. (b-d) are \methodname{} (w/o TRT) on our YouTube VIS sub-train/sub-val split ((b)\&(d) without COCO joint training). We highlight our design choices in gray.
  \vspace{-2pt}
	  }
  \label{table:ablation-design}
\end{table*}

\vspace{5pt}
\noindent\textbf{Datasets.} YouTube VIS~\cite{yang2019video} is a video instance segmentation dataset for detection, segmentation, and tracking of object instances in videos. It contains 2883 high-resolution YouTube videos of 40 common objects such as person, animals, and vehicles, at a frame rate of 30 FPS. The train, validation, and test set contain 2238, 302, and 343 videos, respectively. Every 5th frame of each video is annotated with pixel-level instance segmentation ground-truth masks.  Since we only perform instance segmentation (without tracking), we cannot directly use the validation server of YouTube VIS to evaluate our method. Instead, we further divide the training split into two train-val splits with a 85\%-15\% ratio (1904 and 334 videos). To demonstrate the validity of our own train-val split, we created two more splits, and configured them so that any two splits have video overlap of less than 18\%. We evaluated Mask R-CNN, YOLACT, and \methodname{} on all three splits, the AP variance is within $\pm 2.0$. 

We also evaluate our approach on the MS COCO~\cite{lin2014microsoft} dataset, which is an image instance segmentation benchmark, using the standard metrics. We train on the train2017 set and evaluate on the val2017 and test-dev sets.

\subsection{Instance Segmentation Results}

We first compare \methodname{} to state-of-the-art real-time methods on YouTube VIS using the RTX 2080 Ti GPU in Table~\ref{table:mask-results}.   YOLACT~\cite{bolya2019yolact} with a R101 backbone produces the highest box detection and instance segmentation accuracy over all competing methods.  Our approach, YolactEdge, offers competitive accuracy to YOLACT, while running at a much faster speed (177.6 FPS on a R50 backbone). Even without the TensorRT optimization, it still achieves over 60 FPS for both R50 and R101 backbones, demonstrating the contribution of our partial feature transform design which allows the model to skip a large amount of redundant computation in video.

In terms of mask quality, because YOLACT/YolactEdge produce a final mask of size 138x138 directly from the feature maps without repooling (which potentially misalign the features), their masks for large objects are noticeably higher quality than Mask R-CNN.  For instance, in Fig.~\ref{fig:mask-quality}, both YOLACT and \methodname{} produce masks that follow the boundary of the feet of lizard and zebra, while those of Mask R-CNN have more artifacts. This also explains YOLACT/YolactEdge's stronger quantitative performance over Mask R-CNN on YouTube VIS, which has many large objects. Moreover, our proposed partial feature transform allows the network to take the computed high resolution $C_3$ features to help generate prototypes. In this way, our method is less prone to artifacts brought by misalignment compared to warping all features (as in \cite{zhu2017deep}) and thus can maintain similar accuracy to YOLACT which processes all frames independently. See Fig. \ref{fig:qualitative-results} for more qualitative results.

We next compare \methodname{} to YOLACT on the MS COCO~\cite{lin2014microsoft} dataset in Table~\ref{table:coco_results}. Here YolactEdge is without video optimization since MS COCO is an image dataset.  We compare three backbones: MobileNetv2, ResNet-50, and ResNet-101. Every \methodname{} configuration results in a loss of AP when compared to YOLACT due to the quantization of network parameters performed by TensorRT. This quantization, however, comes at an immense gain of FPS on the Jetson AGX and RTX 2080 Ti. For example, using ResNet-101 as a backbone results in a loss of 0.3 mask mAP from the unquantized model but results in a 20.7/88.3 FPS improvement on the AGX/RTX. We note that the MobileNetv2 backbone has the fastest speed (35.7 FPS on AGX) but has a very low mAP of 20.8 when compared to the other configurations.

Finally, Table~\ref{table:yolactedge_ablation} shows ablations of YolactEdge.  Starting from YOLACT, which is equivalent to YolactEdge without TensorRT and video optimization, we see that with a ResNet-101 backbone, both our video and TensorRT optimizations lead to significant improvements in speed with a bit of degradation in mask/box mAP. The speed improvement for instantiations with a ResNet-50 backbone are not as prominent, because video optimization mainly exploits the redundancy of computation in the backbone stage and its effect diminishes in smaller backbones.

\subsection{Which feature layers should we warp?}
As shown in Table \ref{ablation:partial-feature-transform}, computing $C_3$/$P_3$ features (rows 2-3) yields 1.2-1.4 higher AP than warping $C_3$/$P_3$ features (rows 4-5). We choose to perform partial feature transform over $P$ instead of $C$ features, as there is no obvious difference in accuracy while it is much faster to warp $P$ features.

\subsection{\featflow{}}
To encode pixel motion, \featflow{} takes as input $C_3$ features from the ResNet backbone. As shown in Table \ref{ablation:featflow-reduce-channel}, we choose to reduce the channels to 1/4 before it enters \featflow{} as the AP only drops slightly while being much faster. If we further decrease it to 1/8, the FPS does not increase by a large margin, and flow pre-training does not converge well.
As shown in Table \ref{ablation:featflow-compare-flownets}, accurate flow maps are crucial for transforming features across frames. Notably, our \featflow{} is equally effective for mask prediction as FlowNetS~\cite{dosovitskiy2015flownet}, while being faster as it reuses $C_3$ features for pixel motion estimation (whereas FlowNetS computes flow starting from raw RGB pixels).

\subsection{Temporal Stability}
Finally, although YolactEdge does not perform explicit temporal smoothing, it produces temporally stable masks.\footnote{See supplementary video: \url{https://youtu.be/GBCK9SrcCLM}.} In particular, we observe less mask jittering than YOLACT. We believe this is due to YOLACT only training on static images, whereas YolactEdge utilizes temporal information in videos both during training and testing.  Specifically, when producing prototypes, our partial feature transform implicitly aggregates information from both the previous keyframe and current non-keyframe, and thus ``averages out'' noise to produce stable segmentation masks.

\section{Discussion of Limitations}

Despite YolactEdge's competitiveness, it still falls behind YOLACT in mask mAP. We discuss two potential causes.  

\paragraph{Motion blur} We believe part of the reason lies in the feature transform procedure -- although our partial feature transform corrects certain errors caused by imperfect flow maps (Table~\ref{ablation:partial-feature-transform}), there can still be errors caused by motion blur which lead to mis-localized detections. Specifically, for non-keyframes, $P_4$ and $P_5$ features are derived by transforming features of previous keyframes. It is not guaranteed that the randomly selected keyframes are free from motion blur.  A smart way to select keyframes would be interesting future work.

\paragraph{Mixed-precision conversion} The accuracy gap can also be attributed to mixed precision conversion -- even with the optimal conversion and calibration configuration (Table \ref{table:tensorrt_exhaustive},\ref{ablation:int8-calibration}), the precision gap between training (FP32) and inference (FP16/INT8) is not fully addressed. An interesting direction is to explore \emph{training} with mixed-precision, with which the model could potentially learn to compensate for the precision loss and adapt better during inference. 

\vspace{5pt}
\noindent\textbf{Acknowledgements.}
This work was supported in part by NSF IIS-1751206, IIS-1812850, and AWS ML research award. We thank Joohyung Kim for helpful discussions.

\bibliographystyle{unsrt}
\bibliography{refs}

\end{document}